\documentclass[twocolumn]{article} 
\usepackage{simpleConference}
\usepackage{times}
\usepackage{amssymb}
\usepackage{url,hyperref}

\usepackage{calc}
\newcommand*\rot{\rotatebox{90}}

\usepackage{array}

\usepackage{tikzit}
\usetikzlibrary{shapes.misc, positioning, trees}

\tikzstyle{Node}=[fill=red, draw=black, shape=circle]
\tikzstyle{big}=[draw=black, shape=rectangle, rounded corners, inner sep=6pt]

\tikzstyle{new edge style 0}=[->]

\usepackage{graphicx}
\usepackage{subcaption}

\usepackage{algorithm}
\usepackage{algpseudocode}

\usepackage{color,soul}

\usepackage[export]{adjustbox}
\usepackage[title]{appendix}

\title{Neural Transition-based Parsing of Library Deprecations
}

\author{\hspace{1mm}Petr Babkin\\
	J.P. Morgan AI Research\\
	New York, USA \\
	\texttt{petr.babkin@jpmorgan.com} \\
	\and
	\hspace{1mm}Nacho Navarro \\
	J.P. Morgan AI Research\\
	Madrid, Spain \\
	\texttt{nacho.navarro@jpmorgan.com} \\
	\and
	\hspace{1mm}Salwa Alamir \\
	J.P. Morgan AI Research\\
	London, UK \\
	\texttt{salwa.alamir@jpmorgan.com} \\
	\and
	\hspace{1mm}Sameena Shah \\
	J.P. Morgan AI Research\\
	New York, USA \\
	\texttt{sameena.shah@jpmorgan.com} \\
}

\begin{document}

\maketitle

\begin{abstract}
This paper tackles the challenging problem of automating code updates to fix deprecated API usages of open source libraries by analyzing their release notes.
Our system employs a three-tier architecture: first, a web crawler service retrieves deprecation documentation from the web; then a specially built parser processes those text documents into tree-structured representations; finally, a client IDE plugin locates and fixes identified deprecated usages of libraries in a given codebase\footnote{Demo paper currently under review for ICSE 2023}.
The focus of this paper in particular is the parsing component.
We introduce a novel transition-based parser in two variants: based on a classical feature engineered classifier and a neural tree encoder.
To confirm the effectiveness of our method, we gathered and labeled a set of 426 API deprecations from 7 well-known Python data science libraries, and demonstrated our approach decisively outperforms a non-trivial neural machine translation baseline.
\end{abstract}

\section{Introduction}
\label{introduction}


Modern software projects aim at avoiding the need for reinventing the wheel by relying on reusable functionality provided by open source libraries.
The main downside is that such dependencies can lead to breaking changes in the code when the libraries are updated. In the Python programming language, library contributors use runtime deprecation warnings to alert users of upcoming changes to the API and functionality. However, there is currently no unified method for automatically updating library usages in client code to account for these deprecations, which leads to costly and time consuming manual maintenance without any immediate incentives. This can become especially costly for larger codebases, leading some developers to ignore the deprecations altogether or to lock in to a specific library version \cite{Sawant2019ToRO}. 
Adopting such a strategy could, in the long term, hinder a team's ability to innovate  --- unable to reap the benefits from library's added new features, performance improvements, and bug fixes.



This paper expands on our work reported earlier this year \cite{9794071}, specifically, going in-depth about the development of the parser and its associated novel transition system, as well its re-implementation, from the ground up, in a custom neural architecture.
Additionally, we provide a detailed quantitative and qualitative evaluation against a stronger NMT baseline and close with the discussion of observed limitations and an extensive error analysis.


\section{Related Work}

When it comes to the methodology for processing code and related natural language documentation, various works employ a range of techniques including manual or rule-based labeling of changelogs \cite{nielsen2021semantic}, basic NLP techniques such as TF-IDF and word embeddings e.g, for matching error messages with relevant portions of the documentation \cite{Ansong2021soar},
as well as highly popular end-to-end neural approaches.
The dominant such approach is neural machine translation (NMT), adopted from NLP, which renders various code related tasks such as
\begin{table*}[ht!]
\caption{Identified deprecation categories along with types of workarounds for each (underlined).}
\centering
\begin{tabular}{ | p{4.5em} | p{16em}| p{26em} | } 
 \hline
 Unit & Workaround & Example \\ 
 \hline
 Field & \{none, \underline{other field}\} &  The internal attributes {\em \_start}, {\em \_stop} and {\em \_step} attributes of {\em RangeInde}x have been deprecated. Use the public attributes {\em start}, {\em stop} and {\em step} instead
 \\ 
 \hline
 Method & \{none, \underline{other method}, other method\} &
The {\em Calendar.iterweekdays()} method has been deprecated. Use the {\em Calendar.day\_name} attribute to access the list of weekdays.\\ 
 \hline
 Parameter & \{none, \underline{other parameter}, other method\} &
  Deprecated use of the {\em axis} parameter in the {\em pivot\_table} function. Use the {\em index} and {\em columns} parameters instead.
  \\ 
 \hline
 Namespace & \{none, \underline{module moved}, method moved\} & The {\em urllib} module has been deprecated. Use the {\em urllib.request} module instead. \\ 
 \hline
 Behavior & unsupported & The {\em deprecate()} decorator has been deprecated. Use the {\em warnings} module instead.\\ 
 \hline
\end{tabular}
\label{table:deprecation-types}
\end{table*} 
bug fixing \cite{Jiang2021CURECN}, code summarization \cite{Mastropaolo2021T5} or code review automation \cite{Hellendoorn2021codereview}, as a sequence-to-sequence prediction task using one of the state of the art architectures, such as the encoder-decoder LSTM \cite{deepfix, Chen2021SequenceRSL} or Transformer \cite{Chirkova2021empiricaltransformer}.
Despite the versatility of the seq2seq NLP paradigm, its use for source code generation presents a unique set of challenges, such as syntactic validity and global semantic coherence.
To address some of these challenges, various extensions have been proposed such as a code-aware search strategy \cite{Jiang2021CURECN} and a syntax-aware edit-decoder \cite{Qihao2021editdecoder}.
A promising alternative is to manipulate abstract syntax trees rather than raw lines of code, which has been shown to better capture syntax \cite{Kim2021ast}.
All in all, the level of performance of AI systems for software engineering significantly lags behind that of mainstream NLP, with typical scores ranging between 30\%-50\% across different tasks \cite{deepfix, Hellendoorn2021codereview}.
\section{Approach}

\subsection{Sourcing Library Deprecations}
\label{crawling-deprecations}
Some of the most popular Python libraries use Sphinx\footnote{https://www.sphinx-doc.org/en/master/examples.html} to generate documentation in HTML format that conveniently encloses all code references in special style tags. 
Using this markup allows us to easily identify code entities in the documentation that are subject to deprecations.
We used BeautifulSoup\footnote{https://www.crummy.com/software/BeautifulSoup/bs4/doc/} to extract a list of 410 libraries from the Sphinx projects page, and then queried the Python Package Index API endpoint to obtain a list of versions for each library. 
Next, for each library version, we ran a programmatic Google search to download the actual release notes corresponding web pages. 
Out of the 410 libraries listed on the Sphinx page, we were able to obtain valid results for 154 of them. 
From the release notes pages, we were able to scrape descriptions of individual deprecations by extracting all list items under the "Deprecations" section header.

\subsection{The Structure of Library Deprecations}
\label{annotation-taxonomy}
In order to inform model development, we started by manually analyzing a small
sample of deprecations from the pandas library, with the aim to compile a
systematic taxonomy of different deprecation mechanisms while abstracting away
from the language of individual deprecations. Specifically, it proved useful to
distinguish different units at which a deprecation applies, for example Method,
Parameter, Namespace/Class, Field, and Behavior. Deprecations affecting most of
these types of units involve a change in a method signature or the class's
public interface, whereas Behavior is our catch all category for the internal
changes in the API such as changes in return value types, handling of particular
input types, errors, and side effects. In addition, for each deprecation we
differentiated one of several unit-specific workarounds offered in the
deprecation text, including none. We found these major types to generalize
across multiple release notes. Table \ref{table:deprecation-types} contains
examples of different categories of deprecations we identified.

In addition to the listed workaround types, we have encountered a considerable
number of complicated workarounds involving chaining multiple method calls and
non-trivial parameter algebra, which are left outside of the scope for this
paper.

\subsection{Annotation Requirements for Deprecations}
\label{annotation-requirements}
Having identified the major types of deprecation mechanisms, our next question
is what information needs to be captured from the text of a deprecation to
enable the actual code inspection and update. Provided that the code elements
are given to us for free by the HTML markup, there is no need to perform entity
detection. Given a list of code entities mentioned to in a deprecation, at a
minimum we need to identify which of those elements form deprecated code and
which are part of the workaround. This information alone is not sufficient to
effect the deprecation on a codebase since code entities are connected via
hierarchical relationships, such as parameter-of-method,
as illustrated by the examples. Thus, some form of
relation identification needs to be employed. In addition, further analyses
might need to be carried out, such as entity coreference resolution and working
out the correspondence in cases of multiple code units being deprecated and
replaced simultaneously. While each of these tasks can in principle be solved by
a separate classifier, 
we chose to address them jointly, under the framework of
transition-based parsing, which we describe in the following sections.

\subsection{A Transition-based Parser for Deprecations}
\label{parser}
Transition-based parsing was popularized by Nivre for syntactic dependency
parsing \cite{nivre2008integrating}, and later got extended for constituency and
semantic parsing of abstract meaning representations (AMR) \cite{wang-etal-2015-transition}. Despite
being incremental and essentially greedy, it is competitive with globally
optimal graph-based approaches, while being more computationally efficient and,
with various optimizations, coupled with neural architectures, it achieves state-of-the-art performance across
multiple benchmarks\footnote{http://nlpprogress.com/english/dependency\_parsing.html}. The most basic example of a
transition-based parser is the shift-reduce parser from programming
languages. It operates on two data structures: a buffer (commonly
implemented as a queue) $\beta$ - that holds currently unprocessed input tokens
and a stack $\sigma$ used to store intermediate constituents and the final parse
trees. The core of the parser is a transition system $T$ that defines actions
that change the state of the two data structures, which in case of a shift
reduce parser is comprised of three transitions: $shift$ -- removes the next
element from the queue and pushes it onto the stack; $reduce\_left/right$ (also,
$arc\_left/right$) -- combines the top two elements into a subtree such that the
first/second one is the head (also sometimes called a governor) and the other is
the dependent (defined by the left/right modifier) and place the result onto
the stack. These operations are applied repeatedly until all input elements have
been processed. The shift-reduce transition system (also called arc-standard -- cf.
\cite{nivre2008integrating}) has been formally shown to be complete for deriving
dependency parses that satisfy certain properties such as projectivity \cite{nivre-2008-algorithms}. 
Multiple papers introduced additional transitions to handle issues such as nonprojectivity
and re-entrancy \cite{vilares-gomez-rodriguez-2018-transition, Zhang2016deep}.

Design of the transition system for parsing API deprecations was example-driven, based on manually reviewing a representative sample of deprecations.  
In addition, the choice of a transition system depends on the the type of structures it will be used to derive.
Unlike, syntactic parsing that aims to derive the grammatical structure underlying a natural language text, our goal is to derive a domain-specific semantic representation that can be used in a downstream code analysis component.
This structure is described by the following context free grammar:

\begin{verbatim}
root -> depr [repl]  func -> <code> [arg]
depr -> ns [ns]      arg -> <code>
repl -> ns [ns]      attr -> <code>
                        
ns -> <code> func    ns -> <code>
ns -> <code> attr 
\end{verbatim}
\vspace{0.1cm}

At the top level, denoted as root, the deprecation is comprised of a required deprecated component (denoted as depr) and an optional replacement component repl. 
In turn, the deprecated subtree can consist of one or more namespaces (such as a Python module or a class) that aggregate either an attribute (i.e. a field) or a function, which in turn consist of a code entity (marked up in HTML) and one or more optional arguments, which are also expressed via HTML-marked code entities. 
The replacement subtree mirrors the structure of the deprecation subtree such that there is a one to one correspondence between elements in each.
In addition, a namespace itself can be either deprecated or replaced, in which case it does not contain any nested functions or attributes.

Here is an example deprecation with a corresponding semantic representation, based on the above grammar, in Lisp-style tree notation:

\begin{quote}
Deprecated parameters {\em levels} and {\em codes} in {\em MultiIndex.copy()}.
Use the {\em set\_levels()} and {\em set\_codes()} methods instead.
\end{quote}

\begin{verbatim}
(root
  (depr
    (ns MultiIndex 
      (func copy() (arg levels)))
    (ns MultiIndex 
      (func copy() (arg codes))))
  (repl
    (ns MultiIndex 
      (func set_levels() (arg levels)))
    (ns MultiIndex 
      (func set_codes() (arg codes)))))
\end{verbatim}
\begin{table*}[ht!]
\caption{Derivation of the example parse.}
\label{derivation}
\small\vspace{1mm}
\begin{tabular}{l l p{30em}}
$\tau$           & $\beta$                            & $\sigma$\\ 
\hline
$start$                  & \texttt{{[}levels, ...{]}}            &                                                                                                                                                                                                                                                                                                         \\ \hline
$shift$                & \texttt{{[}codes, ...{]}   }          & \texttt{{[}levels{]}   }                                                                                                                                                                                                                                                                                               \\ \hline
$unary\_x(arg)$          & \texttt{{[}codes, ...{]}  }           & \texttt{{[}(arg levels){]} }                                                                                                                                                                                                                                                                                           \\ \hline
$shift$                & \texttt{{[}MultiIndex.copy, ...{]}} & \texttt{{[}(arg levels), codes{]}   }                                                                                                                                                                                                                                                                                  \\ \hline
$unary\_x(arg)$          & \texttt{{[}MultiIndex.copy, ...{]}} & \texttt{{[}(arg levels), (arg codes){]}   }                                                                                                                                                                                                                                                                            \\ \hline
$shift$                & \texttt{{[}set\_levels, ...{]}}     & \texttt{{[}(arg levels), (arg codes), MultiIndex.copy{]}  }                                                                                                                                                                                                                                              \\ \hline
$reduce\_lx\_each(func)$ & \texttt{{[}set\_levels, ...{]}}     & \begin{minipage}{30em}\small\vspace{1mm}\begin{verbatim}
[(ns MultiIndex (func copy (arg levels))),  
 (ns MultiIndex (func copy (arg codes)))]
\end{verbatim}\end{minipage}\vspace{1mm}                                                                                                                                                               \\ \hline
$reduce\_rx(depr)$       & \texttt{{[}set\_levels, ...{]}}     & \begin{minipage}{30em}\small\vspace{1mm}\begin{verbatim}
[(depr (ns MultiIndex (func copy (arg levels)))
       (ns MultiIndex (func copy (arg codes))))]
\end{verbatim}\end{minipage}\vspace{1mm}                                                                                                                                                              \\ \hline
$shift$                & \texttt{{[}set\_codes{]}  }         & \begin{minipage}{30em}\small\vspace{1mm}\begin{verbatim}
[(depr (ns MultiIndex (func copy (arg levels)))
       (ns MultiIndex (func copy (arg codes)))),   
 set_levels]
\end{verbatim}\end{minipage}\vspace{1mm}                                                                                                                                            \\ \hline
$shift$                &                        & \begin{minipage}{30em}\small\vspace{1mm}\begin{verbatim}
[(depr (ns MultiIndex (func copy (arg levels)))
       (ns MultiIndex (func copy (arg codes)))),   
 set_levels, set_codes]
\end{verbatim}\end{minipage}\vspace{1mm}                                                                                                                          \\ \hline
$reuse\_args\_rx$      &                        & \begin{minipage}{30em}\small\vspace{1mm}\begin{verbatim}
[(depr (ns MultiIndex (func copy (arg levels)))
       (ns MultiIndex (func copy (arg codes)))), 
 (func set_levels (arg levels)), 
 (func set_codes (arg codes))]
\end{verbatim}\end{minipage}\vspace{1mm}                                                                                   \\ \hline
$reuse\_ns\_rx$        &                        & \begin{minipage}{30em}\small\vspace{1mm}\begin{verbatim}
[(depr (ns MultiIndex (func copy (arg levels)))
       (ns MultiIndex (func copy (arg codes)))),
 (ns MultiIndex (func set_levels (arg levels))),
 (ns MultiIndex (func set_codes (arg codes)))]
\end{verbatim}\end{minipage}\vspace{1mm}                                                   \\ \hline
$reduce\_rx(repl)$       &                        & \begin{minipage}{30em}\small\vspace{1mm}\begin{verbatim}
[(depr (ns MultiIndex (func copy (arg levels)))
       (ns MultiIndex (func copy (arg codes)))), 
 (repl (ns MultiIndex (func set_levels (arg levels)))
       (ns MultiIndex (func set_codes (arg codes))))]
\end{verbatim}\end{minipage}\vspace{1mm}                                           \\ \hline
$reduce\_rx(root)$       &                        & \begin{minipage}{30em}\small\vspace{1mm}\begin{verbatim}
[(root 
   (depr (ns MultiIndex (func copy (arg levels)))
         (ns MultiIndex (func copy (arg codes))))
   (repl (ns MultiIndex (func set_levels (arg levels)))
         (ns MultiIndex (func set_codes (arg codes)))))]
\end{verbatim}\end{minipage}\vspace{1mm}\\
\hline
\end{tabular}

\end{table*}

Note in the above example a few crucial things are not overtly expressed but need to be inferred in order to meaningfully capture the deprecation. 
\begin{itemize}
    \item The two referenced parameters are both part of the method copy, however their usage is deprecated independently, and each has a different workaround method. Thus, the method in which the parameter usage is deprecated needs to be duplicated for each parameter to maintain the one-to-one correspondence.
    \item The two replacement methods do not explicitly mention the parameter they are for and this assignment needs to be inferred based on syntactic parallelism of the two sentences.
    \item The class name is mentioned only in the case of the deprecation, but not with the replacement methods. This relationship also needs to be filled in automatically by the parser.
\end{itemize}

First, to adapt the standard shift-reduce transition system to handle constituency and CFG productions, we used the extended transitions introduced by Zhu et al. \cite{zhu-etal-2013-fast}:

$reduce\_rx/lx$ - in contrast to the regular reduce operation, applies a CFG rule to the top two elements of the stack and pushes the resulting constituent on the stack.

$unary\_x$ - applies to the single top element of the stack by raising it to a pre-terminal constituent.

Second, to address the aforementioned issues, we have experimented with multiple alternative transition actions, including those cited in the literature, such as copying, swapping, and various side effects of the existing operations (e.g., copy on reduce). Other alternative solutions we considered involved having multiple stacks for deprecation and replacement subtrees cf. --- \cite{gildea-etal-2018-cache}. 
Our final solution is essentially two-pronged: 1) to introduce a family of "reuse" operations, to systematically account for phenomena of parameter sharing and syntactic parallelism; and 2) to extend the application of the operations beyond the first two elements with certain constraints (e.g., provided all elements are of the same type). 
 
$reduce\_each\_lx$ --- applies the func CFG production iteratively to each of the two arg constituents.

$reuse\_args\_rx$ --- given a fully populated deprecation subtree and two func constituents that will later form the replacement subtree, copies the args of the deprecated funcs onto each of their respective replacement funcs.

$reuse\_ns\_rx$ --- performs a similar copy of the deprecated function namespaces onto the replacement ones.

The complete derivation of the example parse using our transition system is shown in Table \ref{derivation}. Note the effects of applying the extended transitions.

\subsection{Learning The Parser}
\label{classifier}
The transition system described in the previous section merely ensures that the desired semantic representations of code deprecations we have reviewed are in principle derivable through sequential application of correct transitions. The next step is to build a mechanism that would correctly pick transitions on previously unseen texts at inference time. This could be achieved using several different approaches ranging from rules-based to reinforcement learning. We adopt the most common technique that relies on supervised classification of transitions using standard logistic regression. To combat the greediness, we employ beam search decoding with width 10, using the running sum of log-probabilities as the ranking criterion. 

The classifier needs to have good visibility into the parsing configuration as well as the linguistic structure of the sentence. 
For example, to perform a complex reuse operation, the classifier needs to know what the respective types of the constituents being considered and their children are, as well as all of their syntactic relationships in the sentence.
We encode each parsing state into 2 groups of unary features describing the next relevant element of the queue and the stack: $Q_0, S_0$; and 3 groups of interaction features between the queue, the stack, and the top two items on the stack: $Q_0-S_0$, $Q_0-S_1$, $S_0-S_1$. The features in all groups comprise the standard linguistic annotations such as POS tags and dependency paths, obtained from the spaCy library \cite{spacy2} of the code entities and their immediate syntactic parents and children, as well as the semantic labels assigned the constituents parsed so far and their children. Please refer to  Table \ref{table:features} for the complete list. 

\begin{table}[t]
\caption{Transition classifier features. $prefix$ is used to indicate the level of hierarchy in the semantic tree and $i$ is the child index.}
\centering
\begin{tabular}{ | p{9em} | p{4em}| p{8em} | } 
 \hline
 Feature Template & Group & Example Value \\ 
 \hline
\{prefix\}lemma & $Q_0, S_0, S_1$ & $\langle code \rangle$ \\
\{prefix\}dep & & dobj \\
\{prefix\}head & & use \\
\{prefix\}head\_dep & & ROOT \\
\{prefix\}head\_pos & & VERB\\
\{prefix\}root & & deprecate\\
\{prefix\}root\_pos & & VERB \\
\{prefix\}child$_i$ & & method\\
\{prefix\}child\_pos$_i$ & & NOUN\\
\{prefix\}child\_dep$_i$ & & nsubjpass \\
\{prefix\}head\_child$_i$ & & ' '\\
\{prefix\}head\_child\_pos$_i$ & & SPACE\\
\{prefix\}head\_child\_dep$_i$ & & compound\\
\{prefix\}root\_child$_i$ & & method \\ 
\{prefix\}root\_child\_pos$_i$ & & NOUN \\
\{prefix\}root\_child\_dep$_i$ & & dobj \\
\hline
label & $S_0, S_1$ & repl \\
child\_label$_i$ & & ns \\
sub\_child\_label$_i$ & & func \\
\hline
\{prefix\}path\_lemma & $Q_0-S_0$, & $\langle x \rangle$ of method $\langle y \rangle$\\
\{prefix\}path\_pos & $Q_0-S_1$, & X, ADP, NOUN\\
\{prefix\}path\_dep & $S_0-S_1$ & prep, pobj, appos \\
\hline
\end{tabular}
\label{table:features}
\end{table}


\subsection{Neural Representation of the Parser State}
The presented classifier has a number of limitations.
First it requires feature engineering and relies on linguistic annotations such as syntactic dependencies.
Second, discrete representation of features results in high dimensionality and sparseness, which makes it harder to learn a good model, especially with a small dataset.
Third, the model has no visibility into the features of the code symbols it manipulates, ignoring much of possibly relevant information such as morphological e.g., the presence of parentheses could indicate that code denotes a method, or even lexical --- two code entities could be related if they share a common root.
The final limitation is the locality of parser state representation i.e. the model is only able to see the last two items on the stack and queue, but not the global configuration.

These limitations prompted the development of a neural-based parsing model, inspired by Stack-LSTM \cite{Dyer2015TransitionBasedDP} and recursive neural network models such as SPINN \cite{Bowman2016AFU}. 
The main idea is to compute global continuous dense representations of both the queue and the stack.
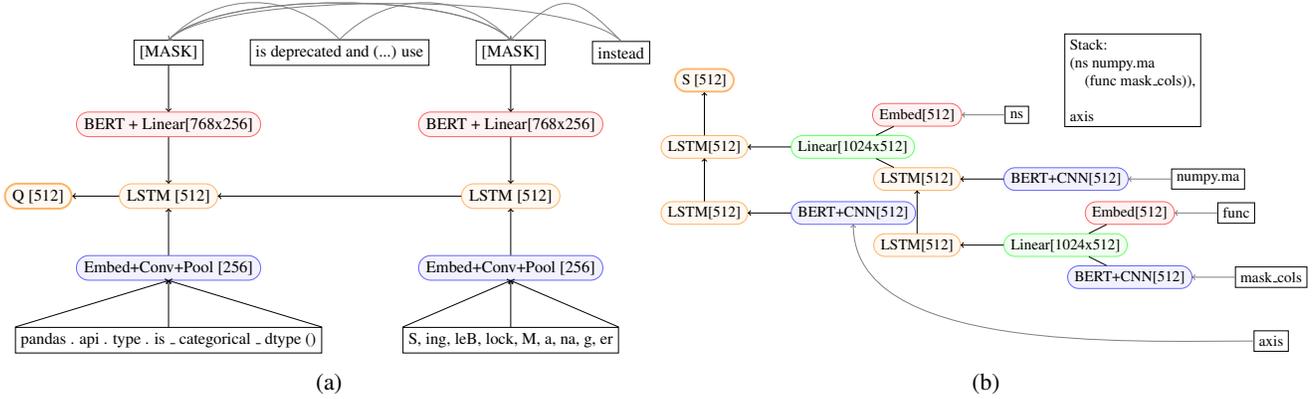
\begin{figure*}
    \begin{subfigure}[b]{0.495\textwidth}
    \centering
    \resizebox{\textwidth}{!}{%
    \begin{tikzpicture}[
    squabluenode/.style={rounded rectangle, draw=blue!60, fill=blue!5, minimum size=5mm},
    squarednode/.style={rounded rectangle, draw=red!60, fill=red!5,  minimum size=5mm},
    squagraynode/.style={rounded rectangle, draw=orange!60, fill=orange!5,  minimum size=5mm},
    distance=1.5cm
    ]
    \centering
    \node[draw] (mask1) {[MASK]}; 
    \node[draw] (text1) [right=of mask1] {is deprecated and (...) use};
    \node[draw] (mask2) [right=of text1] {[MASK]};
    \node[draw] (text2) [right=of mask2]{instead};
    \node[squarednode] (bert1) [below=of mask1] {BERT + Linear[768x256]};
    \node[squarednode] (bert2) [below=of mask2] {BERT + Linear[768x256]};
    
    \draw[<-, gray] (mask1.north) to [bend left=45] (text1.north);
    \draw[<-, gray] (mask1.north) to [bend left=45] (mask2.north);
    \draw[<-, gray] (mask1.north) to [bend left=45] (text2.north);
    \draw[->] (mask1.south) -- (bert1.north);
    
    \draw[<-, gray] (mask2.north) to [bend right=45] (text1.north);
    \draw[<-, gray] (mask2.north) to [bend right=45] (mask1.north);
    \draw[<-, gray] (mask2.north) to [bend left=45] (text2.north);
    \draw[->] (mask2.south) -- (bert2.north);
    
    \node[squagraynode] (lin1) [below=of bert1] {LSTM [512]};
    
    \node[squagraynode] (lin2) [below=of bert2] {LSTM [512]};
    \draw[<-] (lin1.east) -- (lin2.west);
    \node[squagraynode, very thick] (lin0) [left=of lin1] {Q [512]};
    \draw[<-] (lin0.east) -- (lin1.west);
    
    \draw[->] (bert1.south) -- (lin1.north);
    \draw[->] (bert2.south) -- (lin2.north);
    
    \node[squabluenode] (cnn1) [below=of lin1] {Embed+Conv+Pool [256]};
    \node[draw] (bpe1) [below=of cnn1] {pandas . api   .   type   .   is   \_   categorical   \_   dtype   ()};
    
    \draw[->] (bpe1.north east) -- (cnn1.south);
    \draw[->] (bpe1.north) -- (cnn1.south);
    \draw[->] (bpe1.north west) -- (cnn1.south);
    
    \node[squabluenode] (cnn2) [below=of lin2] {Embed+Conv+Pool [256]};
    \node[draw] (bpe2) [below=of cnn2] {S, ing, leB, lock, M, a, na, g, er};
    
    \draw[->] (bpe2.north east) -- (cnn2.south);
    \draw[->] (bpe2.north) -- (cnn2.south);
    \draw[->] (bpe2.north west) -- (cnn2.south);
    
    \draw[->] (cnn1.north) -- (lin1.south);
    \draw[->] (cnn2.north) -- (lin2.south);
    \end{tikzpicture}
    }
    \caption{}
    \label{Q-net}
\end{subfigure}
\begin{subfigure}[b]{0.495\textwidth}
    \centering
    \resizebox{\textwidth}{!}{%
    \begin{tikzpicture}[
    squabluenode/.style={rounded rectangle, draw=blue!60, fill=blue!5, minimum size=5mm},
    squarednode/.style={rounded rectangle, draw=red!60, fill=red!5, minimum size=5mm},
    squagreennode/.style={rounded rectangle, draw=green!60, fill=green!5, minimum size=5mm},
    squagraynode/.style={rounded rectangle, draw=orange!60, fill=orange!5, minimum size=5mm},
    distance=3cm,
    level distance=1.5cm,
      level 1/.style={sibling distance=1.5cm},
      level 2/.style={sibling distance=1.5cm}
    ]
    \centering
    
    \node[squagraynode, very thick] (lstm3)  {S [512]};
    \node[squagraynode] (lstm) [below=of lstm3] {LSTM[512]};
    \node[squagraynode] (lstm0) [below=of lstm] {LSTM[512]};
    
    \draw[->] (lstm.north) -- (lstm3.south);
    \draw[->] (lstm0.north) -- (lstm.south);
    
    \node [squagreennode] (func0) [right=of lstm]{Linear[1024x512]} [grow=right]
        child  {node [squagraynode] (lstm1) {LSTM[512]}}
        child  {node [squarednode] (ns) {Embed[512]}};
        
    \draw[->] (func0.west) -- (lstm.east);

    \node [squabluenode] (func1) [right=of lstm1] {BERT+CNN[512]};
    
    \draw[->] (func1.west) -- (lstm1.east);
        
    \node[squagraynode] (lstm2) [below=of lstm1] {LSTM[512]};
    \draw[<-] (lstm1.south) -- (lstm2.north);
    
      \node [squagreennode] (node2) [right=of lstm2] {Linear[1024x512]} [grow=right]
      child {node [squabluenode] (func2) {BERT+CNN[512]}}
      child {node [squarednode] (fn) {Embed[512]}};
      
      \draw[->] (node2.west) -- (lstm2.east);

    \node [squabluenode] (arg1) [right=of lstm0] {BERT+CNN[512]};
    \draw[->] (arg1.west) -- (lstm0.east);
    
    \node[draw]  (ns1) [right=of ns] {ns};
    \node[draw]  (ma) [right=of func1] {numpy.ma};
    \node[draw]  (fn1) [right=of fn] {func};
    \node[draw]  (maskcols) [right=of func2] {mask\_cols};
    \node[draw]  (axis) [below=of maskcols] {axis};
    
    \draw[->, gray] (ns1.west) -- (ns.east);
    \draw[->, gray] (ma.west) -- (func1.east);
    \draw[->, gray] (fn1.west) -- (fn.east);
    \draw[->, gray] (maskcols.west) -- (func2.east);
    \draw[->, gray] (axis.west) to [in=-90, out=180] (arg1.south);
    
    \node[draw,align=left] at (10,0) {
    Stack:\\
    (ns numpy.ma \\
    \hspace{1em}(func mask\_cols)),\\ 
    \\
    axis};
    \end{tikzpicture}
    }

    \caption{}
    \label{S-net}
\end{subfigure}
    \caption{Neural networks for representing the parser's queue (a) and stack (b). a) Code token's contextual embeddings are concatenated with subtoken CNN representations and fed into an LSTM in reverse order.
    b) For a single code entity \texttt{axis}, the representation is simply its embedding. For a partial tree \texttt{(ns numpy.ma (func mask\_cols))}, the representation is computed recursively from the representations of the tree's children. Each item's representation is then fed into an LSTM.}
\end{figure*}
For the queue, this is accomplished by applying an LSTM over the elements of the queue to produce a fixed size summary of the queue's contents.
\begin{equation}
Q = LSTM(Embed(X))
\end{equation}
The LSTM is reversed so that the representation of the front element can encode information of all elements that follow it in the queue.
The input at each step is the concatenation of two neural representation of the code entity.
To capture the code entity's syntactic context we extract pre-trained BERT embedding \cite{Devlin2019BERTPO} corresponding to the mask token that the code entity is replaced with in the sentence.
We are intentionally making BERT oblivious of the actual code content to focus on its syntactic role in the sentence and to avoid possible confusion between code and natural language.
We empirically verified that tokens predicted by BERT adequately fulfilled the syntactic roles the masked code entities had in the original sentence.
Specifically, it would replace them with nouns like ``text'', ``output'', ``memory'', and ``parameters''.
To reduce the overall number of trainable parameters in the network, we additionally project BERT embeddings to lower dimensionality.
In order to capture morphological and lexical features of code we apply 1D convolutions to the learned byte-pair encodings of subword tokens, and max-pool them into a fixed size representation.
\begin{equation}
Embed(x_i) = [BERT(x_i)W; CNN(x_i)]
\end{equation}

In representing the stack state we similarly apply LSTM to handle variable length sequences of items.
Unlike the queue elements, each element of the stack can be either an atomic code entity (represented as before) or a composition of multiple code entities into a tree-structure of a certain type.
\begin{equation}
S = LSTM(\{s|Embed(s) \cup RecNN(s) \})
\end{equation}
To account for this compositionality, we additionally employ a simple recursive tree encoder that takes the representations of the child code entities $S\prime$ and combines them with a tree's type label embedding $L$ to compute the representation of the parent. 
\begin{equation}
RecNN(s) = tanh(W[L; RecNN(LSTM(S\prime))])
\end{equation}

Both networks are schematically shown in Figures \ref{Q-net} and \ref{S-net}.
One additional LSTM is used to keep track of transitions taken by the parser so far.
The three representations are then concatenated to perform a softmax classification of the parser's next action.

\section{Model Evaluation}
\label{evaluation}

\subsection{Data Collection}
We began collecting data by obtaining a list of all libraries that have documentation generated using Sphinx and finding their release note URLs. As previously noted, the list of libraries we are able to use was reduced from 410 to 154 based on the Sphinx, Pypi name compatibility. From this list we selected 7 popular data science libraries and extracted their deprecation points for annotation in our experiments going forward.

In order to evaluate our system, we annotated a total of 426 deprecations from the release notes of the popular data science libraries, which include pandas, matplotlib, numpy, and networkx.
For each library we randomly sampled several releases with notes containing a deprecation section and allocated them to a random team member to annotate.
\begin{table}[ht!]
\caption{A breakdown of annotated deprecations by lib}
\begin{tabular}{cllllll}
\hline
\multicolumn{1}{r}{\textbf{}} & \multicolumn{1}{r}{pandas} & \multicolumn{1}{r}{matplotlib} & \multicolumn{1}{r}{networkx} & \multicolumn{1}{r}{numpy} & \multicolumn{1}{r}{other} \\
\hline
Total              & 226                                 & 103                                     & 44                                    & 38                                 & 11         \\
Gold              & 98                                  & 52                                      & 31                                    & 11                                 & 2      \\
\hline
\end{tabular}
\end{table}
The annotation task required thoroughly understanding the text of each deprecation and expressing them in terms of a) the code expressions that got deprecated (such as a certain method invocation with a particular parameter) and b) the corresponding replacement code expressions (if any). 
In case of multiple expressions being deprecated we listed them as comma delimited, such that there is a one-to-one correspondence among deprecated code expressions and replacements.
In addition, we have indicated the Unit and Workaround labels of each deprecation, some example having multiple labels.

All annotators followed the same set of annotation guidelines, created based on the analysis of the initial 20 deprecations. 
In case the deprecation description by itself was insufficient or ambiguous for establishing the precise code expressions (e.g. a statement such as ``all subclasses of X''), we chose to consult the library's documentation and pull requests, to obtain the most accurate annotations possible even though that information would not be available to our system during evaluation.

\begin{table}
\caption{Annotation breakdown by unit type (additional infrequent categories totaling 30 examples not shown)}
\begin{tabular}{ccccccccc}
\hline
\textbf{}      & Met. & Beh. & Par. & P. type & Field & P. val. & Ns. \\ \hline
Total & 152             & 85                & 70                 & 51               & 42             & 19                & 16                 \\
Gold  & 108             & 6                 & 55                 & 0                & 28             & 0                 & 13                 \\ \hline
\end{tabular}
\end{table}

Each annotation was reviewed by at least 2 people and was marked as Gold if 1) the type of deprecation Unit and Workaround was in scope for our system; and 2) code annotations were clear and correctly reflected the essence of the deprecation. 
The final set used for evaluation consisted of 194 Gold annotations.

\begin{table}
\caption{Annotation breakdown by workaround type (additional infrequent categories not shown)}
\begin{tabular}{lll}
\hline
Workaround          & Total & Gold \\
\hline
none                              & 120   & 102  \\
other method                      & 73    & 52   \\
chain multiple methods            & 28    & 0    \\
other parameter                   & 18    & 15   \\
other method with parameter       & 12    & 8    \\
complex parameter algebra         & 8     & 0    \\
other value                       & 8     & 0    \\
other method with parameter value & 6     & 0    \\
module moved                      & 6     & 6    \\
other field                       & 6     & 5    \\
method moved                      & 5     & 4   \\
\hline
\end{tabular}
\end{table}

\subsection{A Simple Deterministic Baseline}
\label{sphinx-baseline}
Thanks to the standardized HTML markup generated by Sphinx, all code entities in text are easily identified.
A simple heuristic is to split the text by the word ``deprecated'' as in ``A and B are deprecated; use X and Y instead'' and collect code entities falling on each side from it into two groups: deprecated and replacement.
This heuristic is obviously limited to one simple syntactic configuration, nor is it able to capture compositional relations between code entities, such as methods and parameters.
Nonetheless, it is a reasonable baseline for more sophisticated approaches to beat.

\subsection{A Neural Machine Translation Baseline}
For a stronger baseline, we chose to use a popular sequence-to-sequence neural machine translation (NMT) approach \cite{Bahdanau14neuralmachine} to generate pairs of deprecated and replacement code conditioned on the deprecation texts.
Unlike the word-splitting baseline, NMT is capable of generating compositional code e.g., parameter names nested under method names.
On the other hand, given its generative nature, NMT could produce syntactically invalid code or code references not mentioned in the text.
We used generated code as-is, without any post-processing or filtering.

\subsection{Transition Oracle}
In order to train the parser's classifiers, gold transition labels need to be generated from annotated parse trees.
Typically, in dependency parsing a deterministic algorithm is employed to select gold transitions that would produce the desired parse \cite{nivre-2008-algorithms}.
Given the greater complexity of our transition system we opted for a search-based approach.
We implemented a greedy breadth first search procedure with a maximum breadth of 100 parsing configurations and a parse depth limit of 15, using tree overlap score as the heuristic.
Out of 193 gold trees, the oracle was able to find 113 transition sequences achieving the tree overlap score of at least 90\%.
Transition sequences not reaching this threshold were discarded from training as their resulting trees were considered too noisy.
There are many factors for such an incomplete coverage, including issues with the input and the insufficiency of our transition system.
These are explored in the discussion section.
\begin{table*}[ht!]
    \centering
    \caption{Quantitative evaluation results.}
    \label{table:scores}
    \begin{tabular}{l|ll|lll|llll}
        & EM      & Mean & Met.   & Par. & F./Ns. & 
        \multicolumn{1}{l}{pandas} & 
        \multicolumn{1}{l}{matplotlib} & 
        \multicolumn{1}{l}{networkx} & 
        \multicolumn{1}{l}{numpy + oth.}\\
        \hline
        count & 166 & 166 & 99       & 43        & 35 & 71 & 52 & 31 & 13        \\
        \hline
        Baseline (IOU) & 22 & 16.9 & 21.8 & $<$1   & {\bf 18.} & 34.4 & 5.2 & 0 & 8.3\\
        NMT (F1)  & 14 & 18.9 & 22.6 & 13.7 & 12.7     & 22.9     & 17.2     & 11.5   & 13.   \\
        LR Parser (F1)  &31 &  31.3 & {\bf 45.9} & 15.3 & 4.7 & 42.7 & {\bf 21.5} & {\bf 23.7} & 23.9 \\
        NN Parser (F1) &  {\bf 39} &  {\bf 34.3} & 43.7 &  {\bf 28.9} & 8 &  {\bf 50.3} & 21.1 & 22.4 &  {\bf 24.9}\\
        \hline
    \end{tabular}
\end{table*}

\subsection{Experiments}
We conducted the experiments on the Gold dataset, using our implemented parser models as well as the baselines, and the evaluation metrics described below.
As a weak baseline, we used the simple deterministic heuristic from Section \ref{sphinx-baseline} and reported 
1) hard set equality between code expressions extracted and those annotated, and 
2) intersection-over-union (IOU), each for deprecated and for replacement code, separately.
Both metrics were averaged between the deprecation and replacement code expressions (if present), for the ease of comparison.
The motivation behind using this rather lenient metric was to gauge the ``optimistic lower-bound'' on the task, without subjecting the baseline to the a priori infeasible requirement for handling compositionality.

To evaluate the more sophisticated models we employed a more stringent, compositional metric.
First, we converted code annotations into the tree representation, conforming to the parser's grammar, using Python's ast package. 
We then computed a tree match score as follows: 
If the parsed tree matches the gold tree exactly, return the score of 1. 
Otherwise, decompose each tree into its constituent subtrees using nltk's \texttt{Tree.subtrees} method
\footnote{https://www.nltk.org/\_modules/nltk/tree.html\\ The method returns an exhaustive set of progressively nested subtrees making up a tree, e.g. for (a (b (c d)) (e)) it produces 3 subtrees in addition to the original one: (b (c d)), (c d), and (e).
}.
Next, compute precision and recall, weighted by subtree height, between the two sets of subtrees, and finally aggregate them into an F1 score.
Since the NMT baseline produces a string of tokens rather than a parse tree, in order to report the tree overlap scores, we had to convert its outputs into trees the same way we did for the annotations.
Fortunately, this sequence-to-sequence model consistently produced valid python code, with less than 10\% of outputs failing the conversion due to syntax errors.
For all reported results, including the baselines, we have excluded the examples used for model development to allow for a fair comparison.

Training and evaluation of all models was done through 10-fold cross-validation. 
We aggregate test scores across all folds in Table \ref{table:scores}.
For the NMT baseline, we finetuned the open source huggingface implementation of Facebook's BART encoder-decoder architecture \cite{Wolf2019HuggingFacesTS}, \cite{Lewis2020BARTDS} on our dataset for 3 epochs.
The output was set up as a pair of deprecated and replacement code expressions, delimited by the sentence start token.
For the feature engineered parser, we used sklearn's implementation of the LogisticRegression classifier, with default parameters. 
The neural network parser was implemented in PyTorch, and trained by optimizing cross entropy loss for 30 epochs with Adam, using the learning rate of 5e-4 and dropout of 0.25 
on a single Tesla V100 16GB GPU. 
We experimented with different hyperparameter values as well as including a reconstruction loss auxiliary objective to the tree encoder without much success.
Mean performance scores for all models, and baselines are given in Table \ref{table:scores} along with score breakdowns by deprecation type and library.

The weak baseline system produced 22 exact matches with gold code annotations of deprecated and replacement code expressions (including, no replacement).
The baseline achieves the mean of 16.9 average IOU across non-empty deprecated and replacement expressions.

The NMT baseline resulted in considerably fewer exact matches, possibly because the generation was not restricted to the code entities mentioned in the text. 
The mean score is only slightly better than that of the deterministic baseline, however the performance on parameter deprecations is markedly better, likely due to its ability to generate compositional code.

Both parsers substantially outperform the baselines both in terms of the overall score, as well as across libraries and most deprecation types.
Specifically, the feature engineered parser achieves more than twice the baseline score for method deprecations and shows promising performance improvement for parameter deprecations, especially relative to the weak baseline.
We defer the analysis of both parsers' surprisingly subpar performance on Field and Namespace deprecations to Section \ref{discussion}.

\begin{figure*}
    \begin{subfigure}[b]{0.33\textwidth}
        \includegraphics[width=\textwidth]{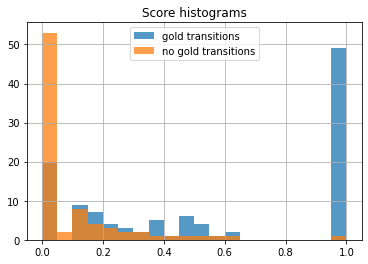}
        \caption{}
        \label{score-histograms}
    \end{subfigure}
    \begin{subfigure}[b]{0.33\textwidth}
        \includegraphics[width=\textwidth]{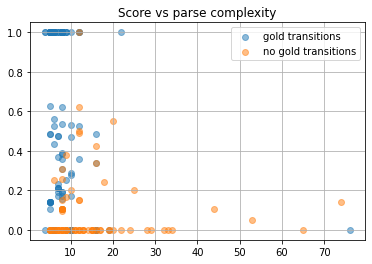}
        \caption{}
        \label{score-complexity}
    \end{subfigure}
    \begin{subfigure}[b]{0.33\textwidth}
        \includegraphics[width=\textwidth]{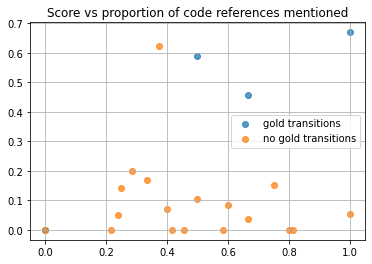}
        \caption{}
        \label{score-mentions}
    \end{subfigure}
    \caption{Quantitative error analysis. (Best viewed in color.)}    
\end{figure*}

The neural parser's results show a moderate further increase in the number of exact matches compared to the feature engineered parser and a slight increase in the mean tree overlap score.
It proved superior at parsing parameter deprecations, nearly doubling the feature engineered parsers's score.
In all other categories, and across different libraries it performed on par with the feature engineered parser.

Overall, the best performance is a achieved on Method deprecations in the pandas library.  
This is in part, because pandas' release notes are generally cleaner with fewer code references, whereas matplotlib and numpy tend to include long lists of deprecations with scarce descriptions.
In the next section we will delve into the qualitative analysis of the results.

\subsection{Discussion}
\label{discussion}
One of the important underlying factors of the quality of predicted parses is the availability of gold transitions.
This becomes apparent when we plot the score histograms for each condition (Figure \ref{score-histograms}).
Highest scoring parses coincided with the trees, for which the oracle succeeded in finding gold transitions.
Conversely, trees for which the oracle failed to find gold transitions also accounted for the majority of the low scoring parses.
Another important consideration that could play a significant role in the parsing result, is the parse complexity.
We operationalized it by simply counting parentheses in the parse's bracketed string representation, which roughly corresponds to the maximum number of nesting levels and siblings in its tree.
The plot on Figure \ref{score-complexity} shows that the majority of highest scoring parses falls under a relatively low ceiling of complexity.
Whereas, partially correct and especially failed parses fall into respectively wider complexity ranges.
Also, consistent with the previous observation, trees without gold transitions tend to result in failed parses regardless of their complexity.
Finally, to test our hypothesis about the effect of code mention coverage on parsing performance we plotted mean scores against the fraction of code references making up a parse tree also mentioned in the text (Figure \ref{score-mentions}).
This visualization confirms that parsing scores positively correlate with the coverage of code entities in the text.

Across the board, the unavailability of gold transitions appears highly predictive of parsing failures.
What made it impossible for the oracle to derive them for such a sizeable fraction of the gold trees? 
A manual review of about 30 examples without gold transitions suggested the chief reasons to be 
a) various issues with the input, such as complete or partial absence of code tags or code entity corruption, 
b) incompleteness of the transition system for deriving especially complex parses,
and c) annotations relying on background knowledge. 
Excluding parse trees not derivable for the above reasons, gives a mean parsing score of about 56\%. 
Among the erroneous parses, 17 contained bona fide confusion errors among different constituent types, such as func vs arg or repl vs depr; 14 parses had either missing or extra constituents; another 17 parses were mostly correct, with the sole discrepancy in the namespace identifier, most of which we found impossible to derive precisely without background knowledge; in addition, 9 deprecation texts contained either duplicate or unrelated code entity mentions, which potentially contributed to spurious predictions.

Zooming into the parser's intrinsic performance, we examine the confusion matrix against the gold transitions (Table \ref{confusion}). 
For most transition types there is a generally high agreement between predicted and true values.
A large portion of the errors are between transitions that are actually  equivalent under certain conditions. 
For example, $reduce\_rx$ when applied to a single item rather than multiple produces the same result as $unary\_x$.
Similarly, the iterative operation $reduce\_lx\_each$ has the same effect as as $reduce\_lx$ if there is only one possible child.
Another major source of errors comes from the consistent overprediction of the $shift$ operation, which is to be expected of the majority class.
In addition, there is some confusion between the argument-less $func$ and $arg$ constituents in the context of the $unary\_x$ operation.
Lastly, some transitions such as $reduce\_lx\_each$ and constituents such as $attr$ could not be effectively learned due to their extremely low frequencies.


Turning to the qualitative results, we are most interested in comparing the strong NMT baseline to the neural network parser.
As pointed out before, NMT is impressive in terms of the validity of the code it outputs even though this is a unrestricted sequence-to-sequence model.
In the examples below we denote NMT output with boldface and the parser's output in a typewriter font.
In the first example, the model correctly picked up the names of the deprecated and replacement functions although also adding some incorrect but not completely implausible namespaces not mentioned in the text.

Input:
\begin{quote}
  The function {\em bellman\_ford} has been deprecated in favor of {\em bellman\_ford\_predecessor\_and\_distance}.
\end{quote}

NMT Output:
\begin{quote}
    \mbox{\texttt{pandas.core.classes.Bellman\_ford()} $\langle{S}\rangle$} \\ \texttt{pl\_toolkits.bellman.ford\_predecessor\\ \_and\_distance()}
\end{quote}

Parser Output:
\begin{verbatim}
(root
  (depr (func bellman_ford))
  (repl (func bellman_ford_predecessor
              _and_distance)))
\end{verbatim}

In some cases, NMT even did a better job than the parser, correctly capturing the compositional relation between the method and its argument while the parser mixed them up.

Input:
\begin{quote}
    The {\em fastpath} keyword in the {\em SingleBlockManager} constructor is deprecated and will be removed in a future version.
\end{quote}

NMT Output:
\begin{quote}
    \texttt{pandas.SingleBlockManager(fastpath)}$\langle{S}\rangle$
\end{quote}

Parser Output:
\begin{verbatim}
(root 
  (depr (func fastpath))
  (repl (func SingleBlockManager)))
\end{verbatim}

On the other hand, in other such simple cases NMT produced completely irrelevant code expressions.

Input:
\begin{quote}
    The {\em squeeze} keyword in {\em groupby()} is deprecated and will be removed in a future version.
\end{quote}

NMT Output:
\begin{quote}
    \texttt{pandas.community.get\_cases()}$\langle{S}\rangle$
\end{quote}

Parser Output:
\begin{verbatim}
(root (depr (func groupby (arg squeeze))))
\end{verbatim}

In more complicated cases, however, the parser clearly showed an advantage, demonstrating a greater capacity at handling multiple code references, and maintaining correct correspondence among the deprecated and replacement ones.

Input:
\begin{quote}
{\em Series.clip\_lower()}, {\em Series.clip\_upper()}, {\em Data-Frame.clip\_lower()} and {\em DataFrame.clip\_upper()} are deprecated and will be removed in a future version.
Use  {\em Series.clip(lower=threshold)}, {\em Series.clip(upper=threshold)} and the equivalent {\em DataFrame} methods.
\end{quote}

NMT Output:
\begin{quote}
    \texttt{Series.clip(), DataFrame.clip \_upper()}$\langle{S}\rangle$
\end{quote}

Parser Output:
\begin{verbatim}
(root
  (depr
    (ns Series (func clip_lower ))
    (ns Series (func clip_upper ))
    (ns DataFrame (func clip_lower ))
    (ns DataFrame (func clip_upper )))
  (repl
    (ns Series 
      (func clip (arg lower=threshold )))
    (ns Series 
      (func clip (arg upper=threshold )))
    (func DataFrame)))
\end{verbatim}

Another complex example:
\begin{quote}
{\em Timestamp.tz\_localize()}, {\em DatetimeIndex.tz\_loca-lize()}, and {\em Series.tz\_localize()} have deprecated the {\em errors} argument in favor of the {\em nonexistent} argument.
\end{quote}

NMT Output:
\begin{quote}
    \texttt{Timestamp.tz\_localize(misc.Datetime- Index.tz(localize)}$\langle{S}\rangle$
\end{quote}

Parser Output:
\begin{verbatim}
(root
  (depr
    (ns Timestamp 
      (func tz_localize (arg errors)))
    (ns DatetimeIndex 
      (func tz_localize (arg errors)))
    (ns Series 
      (func tz_localize (arg errors))))
  (repl
    (ns Timestamp 
      (func tz_localize (arg nonexistent)))
    (ns DatetimeIndex 
      (func tz_localize (arg nonexistent)))
    (ns Series 
      (func tz_localize (arg nonexistent)))
))
\end{verbatim}

\section{Conclusion}
\label{conclusion}
In this paper we presented a novel neural transition-based parsing algorithm for a challenging semantic parsing task.
Both implementations of our parser decisively outperform the strong neural machine translation baseline suggesting there is value in framing the problem as a structured prediction task.
The latter is highly relevant for the field of AI-aided software engineering where model output is subject to stricter requirements than in standard NLP.
Still, there remains a significant performance gap, as in many other SWE-related AI applications, before broad adoption becomes feasible. 
We believe our work can become a stepping stone for future research to help close that gap.

In addition, we contributed an annotated dataset of Python \mbox{library} deprecations and incorporated our algorithm in an IDE \mbox{plugin}.

\paragraph{Disclaimer}
 This paper was prepared for informational purposes by
 the Artificial Intelligence Research group of JPMorgan Chase \& Co\. and its affiliates (``JP Morgan''),
 and is not a product of the Research Department of JP Morgan.
 JP Morgan makes no representation and warranty whatsoever and disclaims all liability,
 for the completeness, accuracy or reliability of the information contained herein.
 This document is not intended as investment research or investment advice, or a recommendation,
 offer or solicitation for the purchase or sale of any security, financial instrument, financial product or service,
 or to be used in any way for evaluating the merits of participating in any transaction,
 and shall not constitute a solicitation under any jurisdiction or to any person,
 if such solicitation under such jurisdiction or to such person would be unlawful.

\bibliographystyle{ACM-Reference-Format}
\bibliography{bibliography}

\begin{appendices}

\begin{table*}[t!]
\scriptsize
\begin{tabular}{rrrrrrrrrrrrrrrrrrrrrrrr}

\textbf{}                       & \rot{\textbf{reduce\_lx(depr)}} & \rot{\textbf{reduce\_lx(func)}} & \rot{\textbf{reduce\_lx(ns)}} & \rot{\textbf{reduce\_lx(root)}} & \rot{\textbf{reduce\_lx\_each(func)}} & \rot{\textbf{reduce\_lx\_each(ns)}} & \rot{\textbf{reduce\_rx(depr)}} & \rot{\textbf{reduce\_rx(func)}} & \rot{\textbf{reduce\_rx(ns)}} & \rot{\textbf{reduce\_rx(repl)}} & \rot{\textbf{reduce\_rx(root)}} & \rot{\textbf{reduce\_rx\_each(func)}} & \rot{\textbf{reuse\_args\_rx()}} & \rot{\textbf{reuse\_funcs\_rx()}} & \rot{\textbf{reuse\_ns\_rx()}} & \rot{\textbf{shift()}} & \rot{\textbf{unary\_x(arg)}} & \rot{\textbf{unary\_x(attr)}} & \rot{\textbf{unary\_x(depr)}} & \rot{\textbf{unary\_x(func)}} & \rot{\textbf{unary\_x(ns)}} & \rot{\textbf{unary\_x(repl)}} & \rot{\textbf{unary\_x(root)}} \\
\textbf{reduce\_lx(depr)}       & 0                         & 0                         & 0                       & 0                         & 0                               & 0                             & 1                         & 0                         & 0                       & 0                         & 0                         & 0                               & 0                          & 0                           & 0                        & 0                & 0                      & 0                       & 0                       & 0                       & 0                     & 0                       & 0                       \\
\textbf{reduce\_lx(func)}       & 0                         & 7                         & 0                       & 0                         & 3                               & 0                             & 0                         & 0                         & 0                       & 0                         & 0                         & 0                               & 0                          & 0                           & 0                        & 0                & 0                      & 0                       & 0                       & 0                       & 0                     & 0                       & 0                       \\
\textbf{reduce\_lx(ns)}         & 0                         & 0                         & 0                       & 0                         & 0                               & 0                             & 0                         & 0                         & 0                       & 0                         & 0                         & 0                               & 1                          & 0                           & 0                        & 0                & 0                      & 0                       & 0                       & 0                       & 0                     & 0                       & 0                       \\
\textbf{reduce\_lx(root)}       & 0                         & 0                         & 0                       & 0                         & 0                               & 0                             & 0                         & 0                         & 0                       & 0                         & 1                         & 0                               & 0                          & 0                           & 0                        & 1                & 0                      & 0                       & 0                       & 0                       & 0                     & 0                       & 0                       \\
\textbf{reduce\_lx\_each(func)} & 0                         & 5                         & 0                       & 0                         & 1                               & 0                             & 0                         & 0                         & 0                       & 0                         & 0                         & 0                               & 0                          & 0                           & 0                        & 0                & 1                      & 0                       & 0                       & 2                       & 0                     & 0                       & 0                       \\
\textbf{reduce\_lx\_each(ns)}   & 0                         & 0                         & 0                       & 0                         & 0                               & 0                             & 0                         & 1                         & 0                       & 0                         & 0                         & 0                               & 0                          & 0                           & 0                        & 0                & 0                      & 0                       & 0                       & 1                       & 0                     & 0                       & 0                       \\
\textbf{reduce\_rx(depr)}       & 0                         & 0                         & 0                       & 0                         & 0                               & 0                             & 70                        & 0                         & 0                       & 1                         & 0                         & 0                               & 0                          & 0                           & 0                        & 8                & 0                      & 0                       & 9                       & 0                       & 0                     & 0                       & 0                       \\
\textbf{reduce\_rx(func)}       & 0                         & 0                         & 0                       & 0                         & 0                               & 0                             & 0                         & 30                        & 0                       & 0                         & 0                         & 3                               & 0                          & 0                           & 0                        & 3                & 0                      & 1                       & 0                       & 12                      & 0                     & 0                       & 0                       \\
\textbf{reduce\_rx(ns)}         & 0                         & 0                         & 0                       & 0                         & 0                               & 0                             & 0                         & 0                         & 0                       & 0                         & 0                         & 0                               & 1                          & 0                           & 0                        & 0                & 0                      & 0                       & 0                       & 4                       & 0                     & 0                       & 0                       \\
\textbf{reduce\_rx(repl)}       & 0                         & 0                         & 0                       & 0                         & 0                               & 0                             & 1                         & 0                         & 0                       & 9                         & 0                         & 0                               & 0                          & 0                           & 1                        & 0                & 0                      & 0                       & 0                       & 0                       & 0                     & 5                       & 0                       \\
\textbf{reduce\_rx(root)}       & 0                         & 0                         & 0                       & 1                         & 0                               & 0                             & 0                         & 0                         & 0                       & 0                         & 64                        & 0                               & 0                          & 0                           & 0                        & 6                & 0                      & 0                       & 0                       & 0                       & 0                     & 0                       & 6                       \\
\textbf{reduce\_rx\_each(func)} & 0                         & 0                         & 0                       & 0                         & 0                               & 0                             & 0                         & 4                         & 0                       & 0                         & 0                         & 6                               & 0                          & 0                           & 0                        & 0                & 0                      & 0                       & 0                       & 0                       & 0                     & 0                       & 0                       \\
\textbf{reuse\_args\_rx()}      & 0                         & 0                         & 0                       & 0                         & 0                               & 0                             & 0                         & 0                         & 0                       & 0                         & 0                         & 0                               & 7                          & 0                           & 0                        & 0                & 0                      & 0                       & 0                       & 1                       & 0                     & 0                       & 0                       \\
\textbf{reuse\_funcs\_rx()}     & 0                         & 0                         & 0                       & 0                         & 0                               & 0                             & 0                         & 0                         & 0                       & 0                         & 0                         & 0                               & 1                          & 18                          & 0                        & 1                & 0                      & 0                       & 0                       & 1                       & 0                     & 0                       & 0                       \\
\textbf{reuse\_ns\_rx()}        & 0                         & 0                         & 0                       & 0                         & 0                               & 0                             & 1                         & 0                         & 0                       & 0                         & 0                         & 0                               & 0                          & 0                           & 13                       & 2                & 0                      & 0                       & 0                       & 0                       & 0                     & 1                       & 0                       \\
\textbf{shift()}                & 0                         & 0                         & 0                       & 0                         & 0                               & 0                             & 4                         & 0                         & 0                       & 1                         & 2                         & 0                               & 0                          & 0                           & 0                        & 235              & 0                      & 0                       & 2                       & 9                       & 0                     & 0                       & 2                       \\
\textbf{unary\_x(arg)}          & 0                         & 1                         & 0                       & 0                         & 1                               & 0                             & 0                         & 3                         & 0                       & 0                         & 0                         & 0                               & 2                          & 0                           & 1                        & 1                & 36                     & 0                       & 0                       & 5                       & 0                     & 0                       & 0                       \\
\textbf{unary\_x(attr)}         & 0                         & 0                         & 0                       & 0                         & 0                               & 0                             & 0                         & 1                         & 0                       & 0                         & 0                         & 0                               & 0                          & 0                           & 0                        & 0                & 0                      & 0                       & 0                       & 0                       & 0                     & 0                       & 0                       \\
\textbf{unary\_x(depr)}         & 0                         & 0                         & 0                       & 0                         & 0                               & 0                             & 14                        & 0                         & 0                       & 0                         & 0                         & 0                               & 0                          & 0                           & 0                        & 5                & 0                      & 0                       & 7                       & 0                       & 0                     & 0                       & 0                       \\
\textbf{unary\_x(func)}         & 0                         & 0                         & 0                       & 0                         & 0                               & 0                             & 0                         & 8                         & 0                       & 0                         & 0                         & 0                               & 0                          & 0                           & 0                        & 2                & 5                      & 0                       & 0                       & 70                      & 2                     & 0                       & 0                       \\
\textbf{unary\_x(ns)}           & 0                         & 0                         & 0                       & 0                         & 0                               & 0                             & 0                         & 0                         & 0                       & 0                         & 0                         & 0                               & 0                          & 0                           & 0                        & 0                & 1                      & 0                       & 0                       & 2                       & 6                     & 0                       & 0                       \\
\textbf{unary\_x(repl)}         & 0                         & 0                         & 0                       & 0                         & 0                               & 0                             & 1                         & 0                         & 0                       & 2                         & 0                         & 0                               & 0                          & 0                           & 0                        & 3                & 0                      & 0                       & 0                       & 0                       & 0                     & 30                      & 0                       \\
\textbf{unary\_x(root)}         & 0                         & 0                         & 0                       & 0                         & 0                               & 0                             & 0                         & 0                         & 0                       & 0                         & 6                         & 0                               & 0                          & 0                           & 0                        & 5                & 0                      & 0                       & 0                       & 0                       & 0                     & 0                       & 24      \\
\end{tabular}
\caption{Confusion matrix (vs oracle transitions)}
\label{confusion}
\end{table*}
\end{appendices}

\end{document}